\documentclass[runningheads]{llncs}
\usepackage{graphicx}
\usepackage{subfig}
\usepackage{wrapfig}
\usepackage{float}
\usepackage{amsmath,amssymb} % define this before the line numbering.
\usepackage{color}
\usepackage[width=122mm,left=12mm,paperwidth=146mm,height=193mm,top=12mm,paperheight=217mm]{geometry}
\usepackage{bbding}
\usepackage{authblk}

\begin{document}
% \linenumbers
\pagestyle{headings}
\mainmatter
\def\ECCV18SubNumber{2289}  % Insert your submission number here

\title{Joint 3D Face Reconstruction and Dense Alignment with Position Map Regression Network} % Replace with your title

\titlerunning{Joint 3D Face Reconstruction and Dense Alignment}

\authorrunning{Y. Feng et al.}

% \author{Yao Feng\thanks{fengyao@sjtu.edu.com}, Fan Wu\thanks{wufan@cloudwalk.cn}, }
\vspace{-3mm}
\author{
	Yao Feng\dag, Fan Wu\ddag, Xiaohu Shao\S, Yanfeng Wang\dag, Xi Zhou\ddag\dag \vspace*{-1mm}\\
% 	{\normalsize \dag Cooperative Madianet Innovation Center, Shanghai Jiao Tong University}\\% \vspace*{-0.8mm}\\
	{\normalsize \dag Shanghai Jiao Tong University, \ddag CloudWalk Technology}\\% \vspace*{-0.8mm}\\
    {\normalsize \S Research Center for Intelligent Security Technology, CIGIT} \\%\vspace*{-1mm}\\
%     {\normalsize \ddag CloudWalk Technology} \vspace*{-1mm}\\
	{\normalsize \texttt{fengyao@sjtu.edu.cn, wufan@cloudwalk.cn, shaoxiaohu@cigit.ac.cn}} \vspace*{-1mm}\\
	{\normalsize \texttt{wangyanfeng@sjtu.edu.cn, zhouxi@cloudwalk.cn}}
\vspace*{-2mm}}
% \author[*]{Author E}
% \affil[*]{Department of Computer Science, \LaTeX\ University}
\institute{\vspace*{-4mm}}

\maketitle
\vspace{-2mm}

\begin{abstract} % model-free

We propose a straightforward method that simultaneously reconstructs the 3D facial structure and provides dense alignment. 
To achieve this, we design a 2D representation called UV position map which records the 3D shape of a complete face in UV space, then train a simple Convolutional Neural Network to regress it from a single 2D image. We also integrate a weight mask into the loss function during training to improve the performance of the network. 
Our method does not rely on any prior face model, and can reconstruct full facial geometry along with semantic meaning.
Meanwhile, our network is very light-weighted and spends only 9.8ms to process an image, which is extremely faster than previous works.
Experiments on multiple challenging datasets show that our method surpasses other state-of-the-art methods on both reconstruction and alignment tasks by a large margin.
Code is available at \url{https://github.com/YadiraF/PRNet}.
\keywords{3D Face Reconstruction, Face Alignment, Dense Correspondence}
\end{abstract}

%%%%%%%%%%%%%%%%%%%%%%%%%%%%%%%%%%%%%%%%%%%%%%%%%%%%%%%%%%%%%%%
\section{Introduction}

% 1. proposed our topic: face reconstruction & face alignment 
3D face reconstruction and face alignment are two fundamental and highly related topics in computer vision. 
In the last decades, researches in these two fields benefit each other.
%they have influenced and promoted mutually.
% 2D alignment 
In the beginning, face alignment that aims at detecting a special 2D fiducial points \cite{zhou2013extensive,zhang2014facial,liang2015unconstrained,peng2016recurrent}
is commonly used as a prerequisite for other facial tasks such as face recognition \cite{wagner2012toward} and assists 3D face reconstruction \cite{zhu2015high-fidelity,Huber2016A} to a great extent.
%--> 3D reconstruction 
However, researchers find that 2D alignment has difficulties \cite{zhao2016fast,jeni2016first} in dealing with problems of large poses or occlusions.
%Meanwhile, many computer vision problems have been well solved by utilizing Convolution Neural Networks(CNNs). 
%% shao, 20180314
With the development of deep learning, many computer vision problems have been well solved by utilizing Convolution Neural Networks (CNNs). 
% 3DMM & Template based -> 3D alignment & face reconstruction.
Thus, some works start to use CNNs to estimate the 3D Morphable Model (3DMM) coefficients \cite{Jourabloo2016Large,zhu2016face,Richardson20163D,liu2016joint,Richardson2016Learning,liu2017dense} or 3D model warping functions  \cite{bhagavatula2017faster,Sela2017Unrestricted} to restore the corresponding 3D information from a single 2D facial image, which provides both dense face alignment and 3D face reconstruction results.
%which can then effortlessly solve the tasks of 3D face reconstruction and dense face alignment.
% Restricted-> performance. 3DMM fitting & TPS Warping -> computation time and complexity
\begin{figure} 
\vspace{-3mm}
\centering
\includegraphics[width=\linewidth]{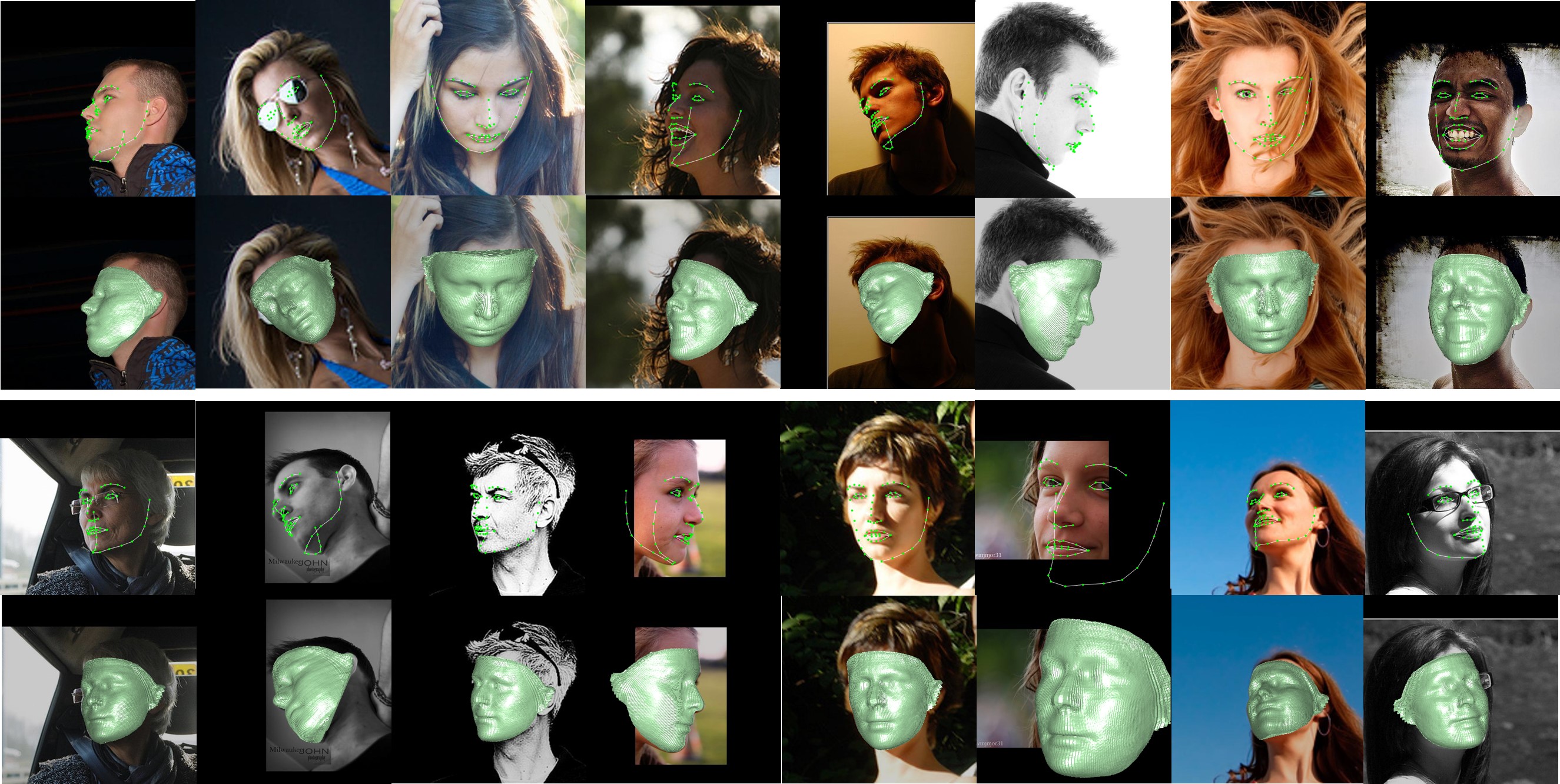} 
\caption{The qualitative results of our method. Odd row: alignment results (only 68 key points are plotted for display). Even row: 3D reconstruction results (reconstructed shapes are rendered with head light for better view).}
\label{fig:examples} %% label for entire figure
\vspace{-5mm}
\end{figure}
However, the performance of these methods is restricted due to the limitation of the 3D space defined by face model basis or templates. The required operations including perspective projection or 3D Thin Plate Spline (TPS) transformation also add complexity to the overall process.
%the computation time and complexity are also negatively effected by perspective projection or 3D Thin Plate Spline(TPS) Transformation. 
% Direct. FAN. VRN

%Recently, two end-to-end works that bypass the limitation of model space achieve state-of-the-art performance on their individual tasks of 3D face reconstruction \cite{Jackson2017Large} and 3D face alignment \cite{Bulat2017How}.
%% 
Recently, two end-to-end works \cite{Jackson2017Large} \cite{Bulat2017How}, which bypass the limitation of model space, achieve the state-of-the-art performances on their respective tasks.
\cite{Bulat2017How} trains a complex network to regress 68 facial landmarks with 2D coordinates from a single image, but needs an extra network to estimate the depth value. Besides, dense alignment is not provided by this method.
\cite{Jackson2017Large} develops a volumetric representation of 3D face and uses a network to regress it from a 2D image. However, this representation discards the semantic meaning of points, 
thus the network needs to regress the whole volume in order to restore the facial shape, which is only part of the volume. So this representation limits the resolution of the recovered shape, and need a complex network to regress it.
%argue
%We surprisingly find that these limitations are not issues in previous model-based methods, which obtains the 3D reconstruction and dense alignment simultaneously. This motivates us to find a method that bypasses all the previous limitations.
% shao, 20180314
We find that these limitations do not exist in previous model-based methods, it motivates us to find a new approach to obtaining the 3D reconstruction and dense alignment simultaneously in a model-free manner.
% 2. Ours
% unrestricted. totally end-to-end, bypassing any 3DMM parameters or TPS function. Fast. 

%In this paper, we propose an end-to-end method that jointly predict the dense alignment and 3D face reconstruction using a simple network.

In this paper, we propose an end-to-end method called Position map Regression Network (PRN) to jointly predict dense alignment and reconstruct 3D face shape. 
% performance
Our method surpasses all other previous works on both 3D face alignment and reconstruction on multiple datasets.
Meanwhile, our method is straightforward with a very light-weighted model which provides the result in one pass with 9.8ms.
%The end-to-end structure and light-weighted network architecture ensure the least computational resources, we can (simultaneously) obtain the 3D reconstruction and dense alignment results (from a single image) in one pass within 9.5ms.
% 3. Details of our work
All of these are achieved by the elaborate design of the 2D representation of 3D facial structure and the corresponding loss function. 
Specifically, we design a UV position map, which is a 2D image recording the 3D coordinates of a complete facial point cloud, and at the same time keeping the semantic meaning at each UV place. 
We then train a simple encoder-decoder network with a weighted loss that focuses more on discriminative region to regress the UV position map from a single 2D facial image. Figure\ref{fig:examples} shows our method is robust to poses, illuminations and occlusions.

In summary, our main contributions are:
% We summarize our contributions as follows:
\begin{itemize}
\item For the first time, we solve the problems of  face alignment and 3D face reconstruction together in an end-to-end fashion without the restriction of low-dimensional solution space.
% \item We demonstrate that face alignment and 3D face reconstruction can be solved together in an end-to-end fashion without the restriction of any morphable or reference model.
\item To directly regress the 3D facial structure and dense alignment, we develop a novel representation called UV position map, which records the position information of 3D face and provides dense correspondence to the semantic meaning of each point on UV space.
\item For training, we proposed a weight mask which assigns different weight to each point on position map and compute a weighted loss. We show that this design helps improving the performance of our network. 
\item We finally provide a light-weighted framework that runs at over 100FPS to directly obtain 3D face reconstruction and alignment result from a single 2D facial image.
% \item We train a light-weighted CNN network which predict the position map from an unconstrained RGB image with a very fast speed.
\item Comparison on the AFLW2000-3D and Florence datasets shows that 
our method achieves more than 25\% relative improvements over other state-of-the-art methods on both tasks of 3D face reconstruction and dense face alignment.
% \item We evaluate our method on multiple datasets of face alignment and 3D face reconstruction. We compare our method with previous works and illustrate that our method outperforms prior work on both tasks with a large margin.
\end{itemize}
%To summarize, our key contributions are:
% \begin{itemize}
% multi-task. end-to-end. model-free. light-weighted. real-time.
%　1. the first time. reconstruct complete facial geometry with its semantic meaning from a single image. end to end. 
%\item To the best of our knowledge, it is the first time to complete the tasks of 3D face reconstruction and 3D face alignment in a model-free manner.
% 2. proposed a representation. 
%\item We propose a new representation that rendering the 3D position of 2D faces to UV space, which can not only represent all the position information of face surface(included non visible region), but also  establish the correspondence between images and 3D template. The position map is demonstrated effective for CNN to learn.
%　3. We complete the regression of position map using CNNs. and design a weight mask to help the location of landmarks.
%\item Utilizing the segmentation information of 3D Morphable Model to generate a weight mask which can be used in the training process to help network focus more on key components (e.g., eyes and mouths).
% 4. Results in Face alignment and Face reconstruction.
%\item In the tasks of 3D face reconstruction from a single image and 3D face alignment, our method achieve a significant improvement over other state-of-the-art methods.
%\item Practical analysis shows the speed of our method, which can run real-time.
%\end{itemize}

%%%%%%%%%%%%%%%%%%%%%%%%%%%%%%%%%%%%%%%%%%%%%%%%%%%%%%%%%%%%%%%%%
\section{Related Works}
As described above, the main problems our method can solve are 3D face reconstruction and face alignment. We will talk about closely related works on these tasks in the following subsections.

\subsection{3D Face Reconstruction}
This part we only talk about 3D face reconstruction from a single image under unconstrained situations. 
% 1. model based
Since Blanz and Vetter proposed 3D Morphable Model(3DMM) in 1999\cite{blanz1999a}, 
methods based on 3DMM are the most popular in completing the task of monocular 3D facial shape reconstruction.
% traditional
Most of earlier methods are to establish the correspondences of the special points between input images and the 3D mean geometry including landmarks\cite{Lee2012Single,zhu2015high-fidelity,Thies2016Face2Face,Huber2016A,Cao2014Displaced,jeni2015dense,grewe2016fully} and local features\cite{Huber2015Fitting,Romdhani2005Estimating,grewe2016fully}, then solve the non-linear optimization function to regress the 3DMM coefficients. 
However, these methods heavily rely on the high accuracy of landmarks or other feature points detector.
% dense correspondence
Thus, some methods\cite{guler2017densereg,Yu2017Learning} firstly use CNNs to learn the dense correspondence between input image and 3D template, then calculate the 3DMM parameters with predicted dense constrains.
% CNN supervised
Recent works also explore the usage of CNN to predict 3DMM parameters directly. 
\cite{Jourabloo2016Large,zhu2016face,Richardson20163D,liu2016joint,Richardson2016Learning} use cascaded CNN structure to regress the accurate 3DMM coefficients, which take a lot of time due to iterations.
\cite{Dou2017End,Tran2016Regressing,jourabloo2015pose,laine2016facial} propose end-to-end CNN architectures to directly estimate the 3DMM shape parameters.
% CNN unsupervised
Unsupervised methods have been also researched recently, \cite{Tewari2017MoFA,Bas20173D} can regress the 3DMM coefficients without the help of training data by using facial texture information, which performs badly in faces with large-poses and unbalanced illuminations.
However, the main defect of those methods are model-based, resulting in a limited geometry which is constrained in model shape space and post-processing to generate 3D mesh from estimated parameters.
% 2. template based
Some other methods can reconstruct 3D faces without 3D shape basis, while still rely on a 3D facial template.
\cite{Hassner2013Viewing,Kemelmacher20113D,gu20063d,Sela2017Unrestricted,santa20163d} can produce a 3D structure by warping the shape of a reference 3D model.
\cite{bhagavatula2017faster} also reconstruct the 3D shape of faces by learning a 3D Thin Plate Spline(TPS) warping function via a deep network which warps a generic 3D model to a subject specific 3D shape. 
Obviously, the reconstructed face geometry from these methods are also restricted by the reference model, which means the structure differs when the template changes.
% 3. model free / end to end
Recently, \cite{Jackson2017Large} propose to straightforwardly map the image pixels to full 3D facial structure via volumetric CNN regression. This method is not restricted in the model space any more, While needs a complex network structure and a lot of time to predict the voxel data.
% 4. ours
Different from above methods, Our framework is model-free and light-weighted, can run at real time and directly obtain the full 3D facial geometry along with its correspondence information. 

% fine details
% \cite{Richardson2016Learning} reconstruct detailed geometric structure by a FineNet, 

%--------------------------------------------------------------
\subsection{Face Alignment}
In the field of computer vision, face alignment is a long-standing problem which attracts a lot of attention.
% generic: 2D face alignment
In the beginning, there are a number of 2D facial alignment approaches which aim at locating a set of fiducial 2D facial landmarks, such as classic Active Appearance Model(AMM)\cite{matthews2004active,saragih2007nonlinear,tzimiropoulos2013optimization} and Constrained Local Models(CLM)\cite{Kim2013Deformable,asthana2013robust}. 
Then cascaded regression\cite{dollar2010cascaded,Xiong2015Global} and CNN-based methods\cite{liang2015unconstrained,peng2016recurrent,Bulat2017How} are largely used to achieve state-of-the-art performance in 2D landmarks location. 
% ? 
However, 2D landmarks location only regresses visible points on faces, which leads to difficulties and is limited to describe face shape when the pose of faces is large.
% recently: 3D face alignment
Recent works then research the 3D facial alignment, which begins with fitting a 3DMM\cite{mcdonagh2016joint,zhu2016face,gou2016shape} or registering a 3D facial template\cite{santa20163d,de20163d} with a 2D facial image. Obviously, 3D reconstruction methods based on model can easily complete the task of 3D face alignment by selecting x,y coordinates of landmarks vertices in reconstructed geometry. Actually, \cite{zhu2016face,Yu2017Learning,jourabloo2015pose} are specially designated methods to achieve 3D face alignment by means of 3DMM fitting. 
Recently \cite{bulat2016two,Bulat2017How} use a deep network to directly predict the heat map to obtain the 3D facial landmarks and achieves state-of-the-art performance. 
% Dense face alignment(3DMM)
Thus, as sparse face alignment tasks are highly completed by aforementioned methods, the task of dense face alignment begins to develop.
Notice that, the dense face alignment means the methods should offer the correspondence between two face images as well as between a 2D facial image and a 3D facial reference geometry. \cite{liu2017dense} use multi-constraints to train a CNN which estimates the 3DMM parameters and then provides a very dense 3D alignment.
% Furthermore, all the methods mentioned above can not infer the correspondence of invisible pixels on input image without post-hoc 3D fitting. 
\cite{guler2017densereg,Yu2017Learning} directly learn the correspondence between a 2D input image and a 3D template via a deep network, while those predicted correspondence is not complete, only visible face-region is considered.
Compared to prior works, our method can directly establish the dense correspondence between all region in faces and 3D template once the position map is regressed.
No intermediate parameters such as 3DMM coefficients and TPS warping parameters are needed in our method, which means our network can run very fast bypassing complex operations including perspective projection and TPS transformation.
% ours

%------------------------------------------------------
% \subsection{UV Texture} 
% \cite{bhagavatula2017faster,maninchedda2016face} represent 3D face as 2.5D height map which stored in a cylindrical projection of 3D model.

%%%%%%%%%%%%%%%%%%%%%%%%%%%%%%%%%%%%%%%%%%%%%%%%%%%%%%%%%%%%%%%%%%%%%%%%%%%%%%%%%   
%\section{Face Position Map Regression}
% \section{Regression of Facial Position Map}
\section{Proposed Method}
% \section{Position map Regression Network (PRN)}
This section describes the framework and the details of our proposed method. Firstly, we introduce the characteristics of the position map for our 3D face representation. Then we elaborate 
the CNN architecture and the loss function designed specially for learning the mapping from unconstrained RGB image to its 3D structure. The implementation details of our method are shown in the last subsection.

\subsection{3D Face Representation}
\label{sec: representation}
Our goal is to regress the 3D facial geometry and its dense correspondence information from a single 2D image. Thus we need a proper representation which can be directly predicted via a deep network.
% However, this is difficult to regress since projection from 3D space into 1D vector discards the adjacency information among points, while adjacent points in space usually could share network weights in predicting their coordinates, thus a network needs a fully connected layer with a lot of weight parameters to predict the coordinates as a 1D vector. 
% 1D
One simple and commonly used idea is to concatenate the coordinates of all points in 3D face as a vector and use a network to predict it. However, this transformation increases the difficulties in training since projection from 3D space into 1D vector discards the spatial adjacency information among points. 
While it's natural to think that spatially adjacent points could share weights in predicting their positions, which can be easily achieved by using convolutional layers. The coordinates as a 1D vector needs a fully connected layer to predict each point with much more parameters, which increases the network size and is hard to train.
%In the learning process of networks, adjacent points in space usually could share network weights which can be easily achieved by convolutional layers. While the coordinates as a 1D vector needs a fully connected layer with a lot of weight parameters to predict, which is more difficult to learn.  
\cite{Fan2016A} proposed a point set generation network to directly predict the point cloud of 3D object as a vector from a single image. However, the max number of points is only 1024, far from enough to represent an accurate 3D face. 
% 1D 3DMM 
% So model-based methods\cite{zhu2016face,Dou2017End,liu2017dense} learn a linear basis of the 1D space and regress the parameters of the linear space instead, but this inevitably limit the resolution of the predicted face.   
So model-based methods\cite{zhu2016face,Dou2017End,liu2017dense} regress a few model parameters rather than the coordinates of points, which usually needs special care in training such as using Mahalanobis distance and inevitably limits the estimated face geometry to the their model space.   
% 3D 
\cite{Jackson2017Large} proposed 3D binary volume as the representation of 3D structure and uses Volumetric Regression Network(VRN) to output a $192 \times 192 \times 200$ volume as the discretized version of point cloud. By using this representation, VRN can be built with full convolutional layers. However, discretization limits the resolution of point cloud, and most part of the network output correspond to non-surface points which are of less usage.

\begin{figure}
\vspace{-3mm}
\centering
\includegraphics[width=0.9\textwidth]{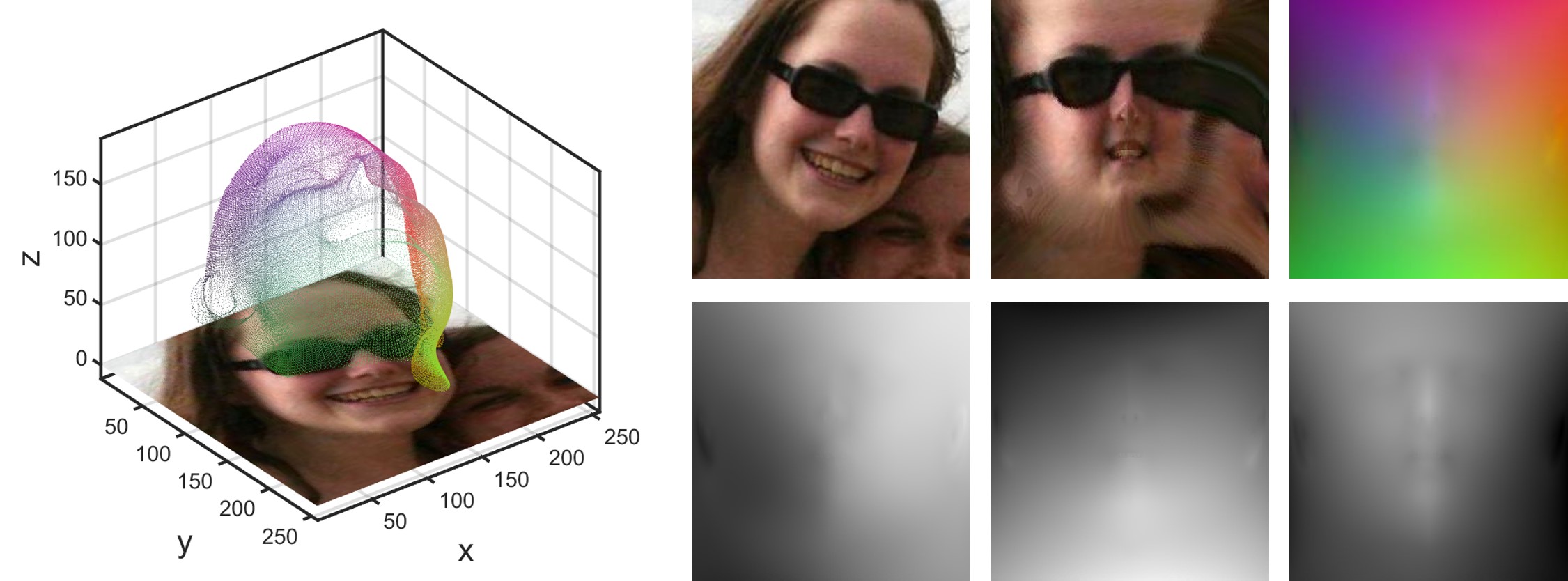} 
\caption{The illustration of UV position map. Left: 3D plot of input image and its ground truth 3D point cloud. Right: The first row is the input 2D image, extracted UV texture map and corresponding UV position map. The second row is the x, y, z channel of the UV position map.}
\label{fig: map}
\vspace{-3mm}
\end{figure}

% UV Position Map
To address the problems in previous works, we propose UV position map as the presentation of full 3D facial structure.
UV position map or position map for short, is a 2D image recording 3D positions of all points in UV space. In the past years, UV space or UV coordinates, which is a 2D image plane parameterized from the 3D space, has been utilized as a way to express information including the texture of faces(texture map)
\cite{Bas20173D,deng2017uv,moschoglou2017multi,xue2018side}, 2.5D geometry(height map)\cite{maninchedda2016face,maninchedda2017fast} and the correspondences between 3D facial meshes\cite{booth2014optimal}.
Different from previous works, we use UV space to store the 3D coordinates of points from 3D face model. 
% x,y,z
As shown in Figure \ref{fig: map}, we define the 3D face point cloud in Left-handed Cartesian Coordinate system. The origin of the 3D space overlaps with the upper-left of the input image, with the positive x-axis pointing to the right of the image. The ground truth 3D face point cloud exactly matches the face in the 2D image when projected to the x-y plane. Thus our position map can be easily comprehended as replacing the $r$, $g$, $b$ value in texture map by $x$, $y$, $z$ coordinates.

% correspondence
% Then unconstrained 2D facial images and their corresponding 3D shapes are required to form our training 
In order to keep the semantic meaning of points within position map, we create our UV coordinates based on 3DMM.
Since we want to regress the 3D full structure directly, the unconstrained 2D facial images and their corresponding 3D shapes are needed for end-to-end training. 300W-LP\cite{zhu2016face} is a large dataset that contains
more than 60K unconstrained images with fitted 3DMM parameters, which is suitable to form our training pairs. Besides, the 3DMM parameters of this dataset are based on the Basel Face Model(BFM)\cite{blanz1999a}. 
% We choose 300W-LP dataset which is constructed by \cite{zhu2016face} to construct the training pairs. The dataset
% \cite{zhu2016face} conducts a large dataset called 300W-LP that contains unconstrained images with its fitted 3DMM parameters. The Basel Face Model(BFM) is used as the 3D base model, which is the first 3D Morphable Model proposed by \cite{blanz1999a}. 
Thus, in order to make full use of this dataset, we conduct the UV coordinates corresponding to BFM. 
To be specific, we use the parameterized UV coordinates from \cite{Bas20173D} which computes a Tutte embedding\cite{Floater1997Parametrization} with conformal Laplacian weight and then maps the mesh boundary to a square. 
Since the number of vertices in BFM is more than 50K, we choose $256 \time  256$ as the position map size, which get a high precision point cloud with negligible re-sample error.

% Advantages of UV Position Map over other 2D representations
Thus our position map records a dense set of points from 3D face with its semantic meaning, we are able to simultaneously obtain the 3D facial structure and dense alignment result by using a CNN to regress the position map directly from unconstrained 2D images. The network architecture in our method could be greatly simplified due to this convenience.
Notice that the position map contains the information of the whole face, which makes it different from other 2D representations such as Projected Normalized Coordinate Code (PNCC)\cite{zhu2016face,Richardson2016Learning}, an ordinary depth image\cite{Sela2017Unrestricted} or quantized UV coordinates\cite{guler2017densereg}, which only reserve the information of visible face region in the input image. Our proposed position map also infers the invisible parts of face, thus our method can predict a complete 3D face.

%-----------------------------------------------------------------------------------------

\subsection{Network Architecture and Loss Function}

\begin{figure}
\vspace{-3mm}
\centering
\includegraphics[width=0.9\textwidth]{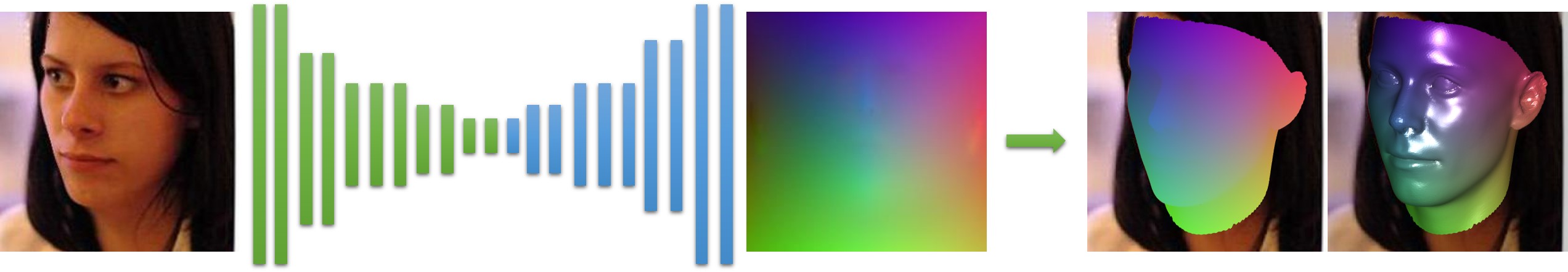} 
\caption{The architecture of PRN. The Green rectangles represent the residual blocks, and the blue ones represent the transposed convolutional layers.}
\label{fig: network architecture}
\vspace{-5mm}
\end{figure}
% network
Since our network transfers the input RGB image into position map image, we employ an encoder-decoder structure to learn the transfer function. 
The encoder part of our network begins with one convolution layer followed by 10 residual blocks\cite{He2016Deep} which reduce the $256 \times 256 \times 3$ input image into $8 \times 8 \times 512$ feature maps, the decoder part contains 17 transposed convolution layers to generate the predicted $256 \times 256 \times 3$ position map. 
We use kernel size of 4 for all convolution or transposed convolution layers, and use ReLU layer for activation.
Given that the position map contains both the full 3D information and dense alignment result, we don't need extra network module for multi-task during training or inferring.
The architecture of our network is shown in Figure \ref{fig: network architecture}. 

% Loss Function
In order to learn the parameters of the network, we build a novel loss function to measure the difference between ground truth position map and the network output. Mean square error (MSE) is a commonly used loss for such learning task, such as in \cite{Yu2017Learning,Crispell2017Pix2face}. However, MSE treats all points equally, so it is not entirely appropriate for learning the position map. Since central region of face has more discriminative features than other regions, we employ a weight mask to form our loss function.

\begin{figure}
\vspace{-3mm}
  \centering
  \subfloat{
    \includegraphics[width=0.2\linewidth]{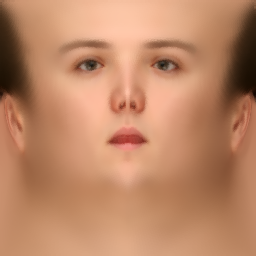}}
%   \hspace{1in}
  \subfloat{
    \includegraphics[width=0.2\linewidth]{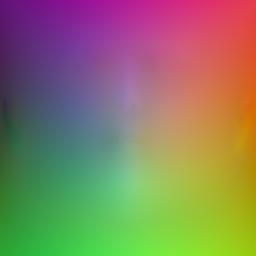}}
      %\hspace{1in}
  \subfloat{
    \includegraphics[width=0.2\linewidth]{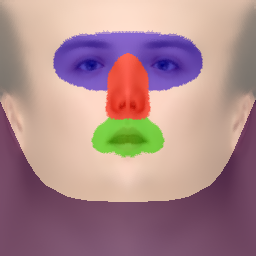}}
  \subfloat{
    \includegraphics[width=0.2\linewidth]{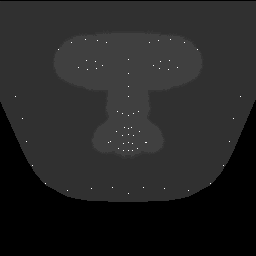}}
\caption{The illustration of weight mask. From left to right: UV texture map, UV position map, colored texture map with segmentation information (blue for eye region, red for nose region, green for mouth region and purple for neck region), the final weight mask.}
\label{fig: weight mask}
\vspace{-5mm}
\end{figure}

As shown in Figure \ref{fig: weight mask}, the weight mask is a gray image recording the weight of each point on position map. It has the same size and pixel-to-pixel correspondence to position map. According to our objective, we separate points into four categories, each has its own weights in the loss function. 
The position of 68 facial keypoints has the highest weight, so that to ensure the network to learn accurate locations of these points.
The neck region usually attracts less attention, and is often occluded by hairs or clothes in unconstrained images. Since learning the 3D shape of neck or clothes is beyond our interests, we assign 0 weight to points in neck region to reduce disturbance to the training process.

Thus, we denote the predicted position map as $P(x,y)$ for $x$, $y$ representing each pixel coordinate. Given the ground truth position map $\tilde{P}(x, y)$ and weight mask $W(x,y)$, our loss function is defined as:
\begin{equation}
Loss = \sum{\lVert P(x, y) - \tilde{P}(x, y) \rVert \cdot W(x, y)}
\end{equation}

Specifically, We use following weight ratio in our experiments, subregion1 (68 facial landmarks): subregion2 (eye, nose, mouth): subregion3 (other face area): subregion4 (neck) = 16:4:3:0. 
The final weight mask is shown in Figure\ref{fig: weight mask}

\subsection{Training Details}

Because our training process needs datasets containing both 2D face images and its corresponding 3D point clouds with semantic meaning, we choose \textit{e.g.}, \textbf{300W-LP}\cite{zhu2016face} to form our training sets, since it contains face images across different angles with the annotation of estimated 3DMM coefficients, from which the 3D point cloud could be easily recovered.
Specifically, we crop the images according the ground truth bounding box and rescale them to size $256 \times 256$. Then utilize their annotated 3DMM parameters to generate the corresponding 3D position, and render them into UV space to obtain the ground truth position map, the map size in our training is also $256 \times 256$, which means a precision of more than 45K point cloud to regress. 
Notice that, although our training data is generated from 3DMM, our network's output, the position map is not restricted to any face template or linear space of 3DMM.

We perturb the training set by randomly rotating and translating the target face in 2D image plane. To be specific, the rotation is from -45 to 45 degree angles, translation changes is random from 10 percent of input size, and scale is from 0.9 to 1.2. 
Like \cite{Jackson2017Large}, we also augment our training data by scaling color channels with scale range from 0.6 to 1.4.
In order to handle images with occlusions, we synthesize occlusions by adding noise texture into raw images, which is similar to the work of \cite{saito2016real,Yu2017Learning}. 
With all above augmentation operations, our training data covers all the difficult cases.
%The augmented training data covers all the difficult cases, because of the great learning of our designed RPN network, our trained model is able to handle input face images under such challenging situations.

We use the network described \ref{fig: network architecture} to train our transfer model. For optimization,  we use Adam optimizer with a learning rate begins at 0.0001 and decays half after each 5 epochs. The batch size is set as 16. All of our training codes are implemented with TensorFlow\cite{abadi2016tensorflow}.

%%%%%%%%%%%%%%%%%%%%%%%%%%%%%%%%%%%%%%%%%%%%%%%%%%%%%%%%%%%%%%%%%%%%%%%%%%%%%%%%%%%%%
\section{Experimental Results}
\label{sec: exp}

In this part, we evaluate the performance of our proposed method on the tasks of 3D face alignment and 3D face reconstruction. 
We first introduce the test datasets used in our experiments in section \ref{test data}.
Then in section \ref{sec: alignment} and \ref{sec: reconstruction} we compare our results with other methods in both quantitative and qualitative way.
We then compare our method's runtime with other methods in section \ref{sec: runtime}.
In the end, the ablation study is conducted in section \ref{sec: ablation} to evaluate the effect of weight mask in our method.

\subsection{Test Dataset}
\label{test data}
To evaluate our performance on the task of dense alignment and 3D face reconstruction, multiple test datasets listed below are used in our experiments:

\textbf{AFLW2000-3D} 
is constructed by \cite{zhu2016face} to evaluate 3D face alignment on challenging unconstrained images. 
This database contains the first 2000 images from AFLW\cite{koestinger2011annotated} and expands its annotations with fitted 3DMM parameters and 68 3D landmarks. 
We use this database to evaluate the performance of our method on both face reconstruction and face alignment tasks.

\textbf{AFLW-LFPA} % To modify
is another extension of AFLW dataset constructed by \cite{Jourabloo2016Large}. 
By picking images from AFLW according to the poses, the authors construct this dataset which contains 1299 test images with a balanced distribution of yaw angle. 
Besides, each image is annotated with 13 additional landmarks as a expansion to only 21 visible landmarks in AFLW.
This database is evaluated on the task of 3D face alignment. We use 34 visible landmarks as the ground truth to measure the accuracy of our results.

\textbf{Florence}
is a 3D face dataset that contains 53 subjects with its ground truth 3D mesh acquired from a structured-light scanning system\cite{bagdanov2011florence}. On experiments, each subject generates renderings with different poses as the same with \cite{Jackson2017Large}: a pitch of -15,20 and 25 degrees and spaced rotations between -80 and 80. We compare the performance of our method on face reconstruction against other very recent state-of-the-art methods VRN-Guided\cite{Jackson2017Large} and 3DDFA\cite{zhu2016face} on this dataset.

\subsection{3D Face Alignment}
\label{sec: alignment}
\begin{figure}
\vspace{-3mm}
  \centering
  \subfloat{
    \includegraphics[width=0.45\linewidth]{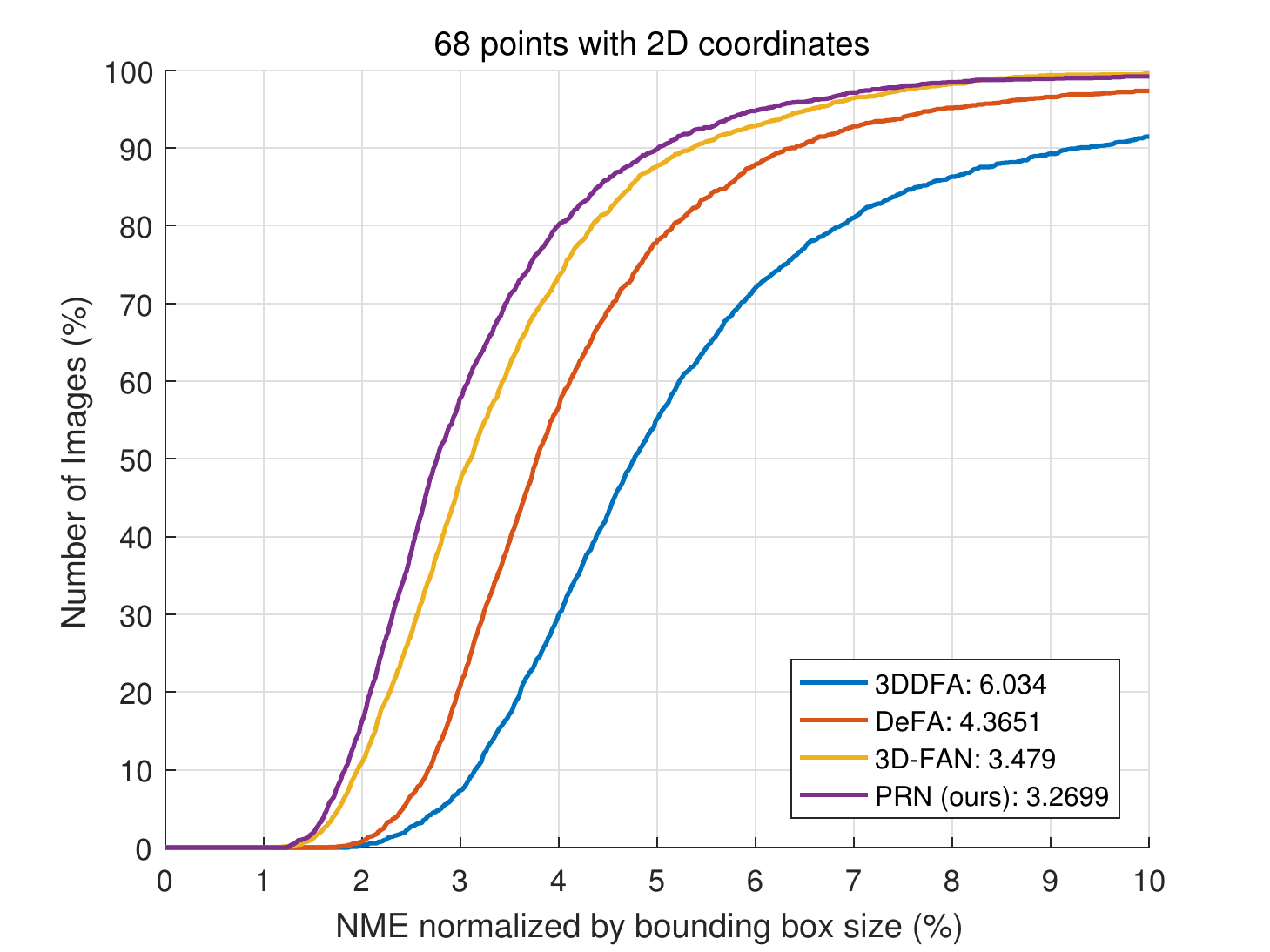}}
  \subfloat{
    \includegraphics[width=0.45\linewidth]{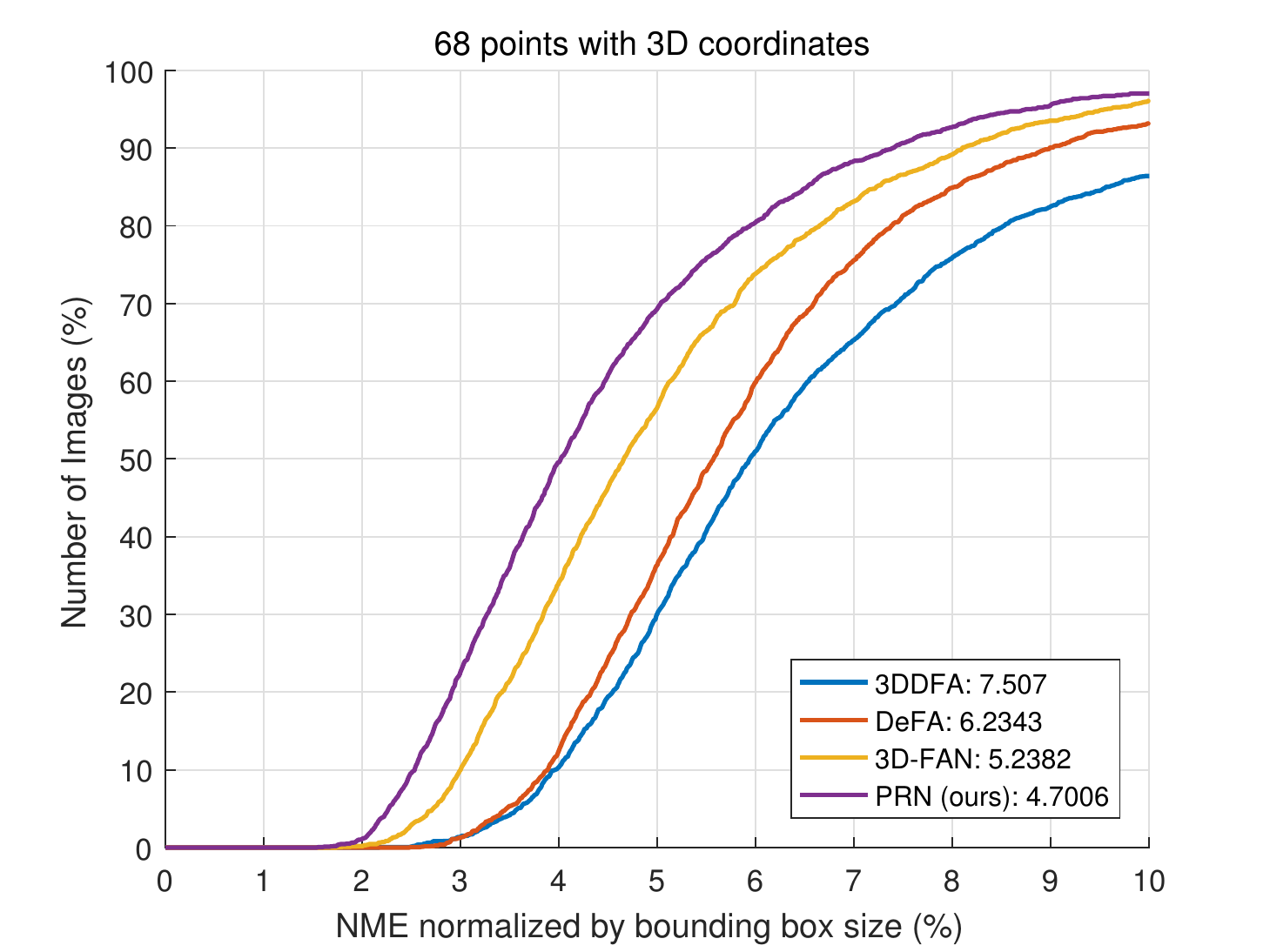}}
  \caption{Cumulative Errors Distribution (CED)
curves on AFLW2000-3D. Evaluation is performed on 68 landmarks with both the 2D(left) and 3D(right) coordinates. Overall 2000 images from AFLW2000-3D dataset are used here. The mean NME\%  of each method is also showed in the legend.}
  \label{fig:aflw2000 68} 
\vspace{-3mm}
\end{figure}

% metric
To evaluate the face alignment performance. We employ the Normalized Mean Error(NME) to be the evaluation metric, bounding box size is used as the normalization factor.
% 68 alignment.(AFLW2000, AFLW)
Firstly, we evaluate our method on a sparse set of 68 facial landmarks, and compare our result with 3DDFA\cite{zhu2016face}, DeFA\cite{liu2017dense} and 3D-FAN\cite{Bulat2017How} on dataset AFLW2000-3D.  
As shown in figure \ref{fig:aflw2000 68}, 
our result slightly outperforms the state-of-the-art method 3D-FAN when calculating per distance with 2D coordinates. When considering the depth value, 
the performance discrepancy between our method and 3D-FAN increases. 
% In both 2D and 3D landmarks localization tasks, our method outperforms.
%the NME of all methods degrades while our method still perform the best and the discrepancy increases. 
Notice that, the 3D-FAN needs another network to predict the z coordinate of landmarks, while the depth value can be obtained directly in our method.

% different poses
To further investigate the performance of our method across poses and datasets, we also report the NME of faces with small, medium and large yaw angles on AFLW2000-3D dataset and the mean NME on both AFLW2000-3D and AFLW-LPFA datasets. 
Table \ref{tab: alignment} shows the comparison with other methods, note that the numerical values are recorded from their published papers except the ones of 3D-FAN. Follow the work \cite{zhu2016face}, we also randomly select 696 faces from AFLW2000 to balance the distribution. 
The result shows that our method is robust to the change of pose and datasets.

\begin{figure} 
\vspace{-3mm}
\centering
\includegraphics[width=\linewidth]{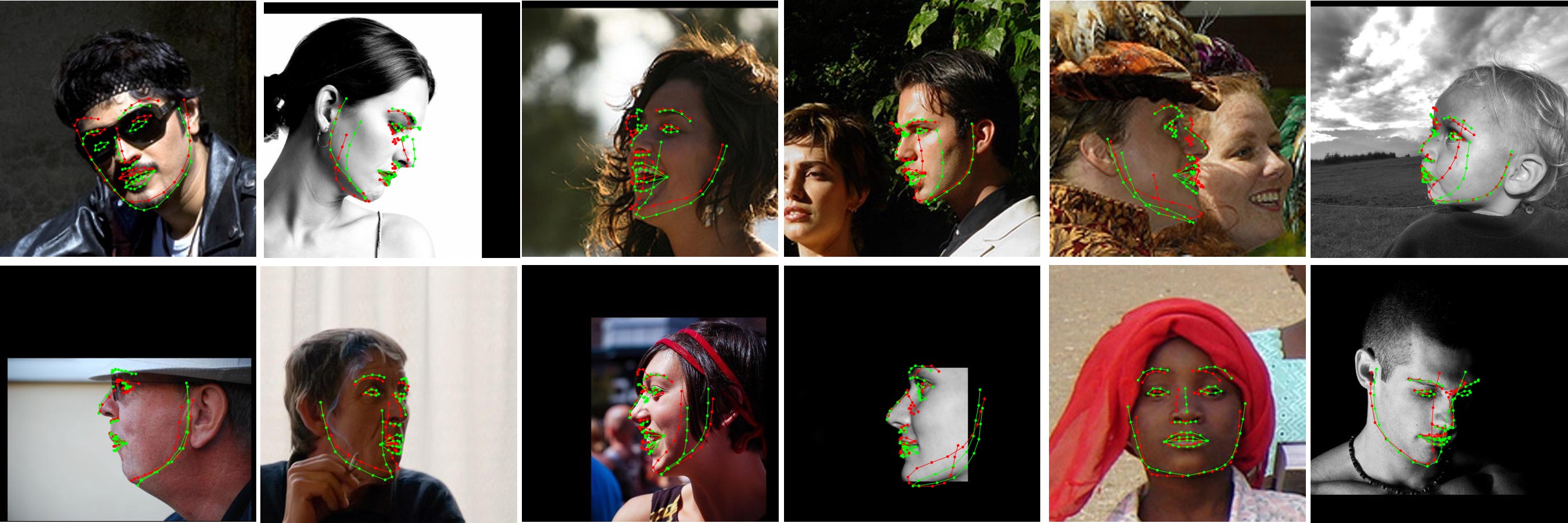} 
\caption{Examples from AFLW2000-3D dataset show that our predictions are more accurate than ground truth in some cases. Green: predicted landmarks by our method. Red: ground truth from \cite{zhu2016face}.}
\label{fig:aflw2000 cases} %% label for entire figure
\vspace{-3mm}
\end{figure}

% qualitative results
Although all the state-of-the-art methods of face alignment conduct evaluation on AFLW2000-3D dataset, the ground truth of this dataset is still controversial\cite{Yu2017Learning,Bulat2017How} due to its annotation pipeline which is based on Landmarks Marching method in \cite{zhu2015high-fidelity}.
Thus, we visualize some results in Figure \ref{fig:aflw2000 cases} that have NME larger than 6.5\% and we find our results are more accurate than the ground truth in some cases.
\begin{table}
%\vspace{-3mm}
\centering
\caption{Performance comparison on two large-pose face alignment datasets AFLW2000-3D(68 landmarks) and AFLW-LFPA(34 visible landmarks). The NME (\%) for faces with different yaw angles are reported. The first best result in each category is highlighted in bold, the lower is the better. 
}
\begin{tabular}{|c|c|c|c|c|c|c|}
   \hline
     & \multicolumn{4}{|c|}{AFLW2000-3D}  & AFLW-LFPA\\
   \hline
   Method & 0 to 30 & 30 to 60 & 60 to 90 & Mean & Mean\\
   \hline
   SDM\cite{Xiong2015Global} & 3.67 & 4.94 & 9.67 & 6.12 & -\\
   \hline
%    PIFA\cite{Xiong2015Global} & - & - & - & - & 6.52\\
   \hline
   3DDFA \cite{zhu2016face} & 3.78 & 4.54 & 7.93 & 5.42 & -\\
   \hline
   3DDFA + SDM \cite{zhu2016face} & 3.43 & 4.24 & 7.17 & 4.94 & -\\
   \hline
   PAWF\cite{Jourabloo2016Large}  & - & - & - & - & 4.72\\
   \hline
   Yu et al. \cite{Yu2017Learning} & 3.62 & 6.06 & 9.56 & - & - \\
   \hline
   3DSTN\cite{bhagavatula2017faster} & 3.15 & 4.33 & 5.98 & 4.49 & - \\
   \hline
   DeFA\cite{liu2017dense} & - & - & - & 4.50 & 3.86\\
\hline
   PRN (ours) & \textbf{2.75} & \textbf{3.51} & \textbf{4.61} & \textbf{3.62} & \textbf{2.93}\\
   \hline
\end{tabular}
\label{tab: alignment}
\vspace{-3mm}
\end{table}

% Figure \ref{fig:aflw2000 cases} shows 10 images both annotated with the ground truth and predicted landmarks by our method. The results indicate that the NME of our performance may be higher if the ground truth was more accurate.
%Some failure examples of our methods are showed in the supplementary materials due to the page limitation.

\begin{figure}
  \vspace{-3mm}
  \centering
  \subfloat{
    \includegraphics[width=0.45\linewidth]{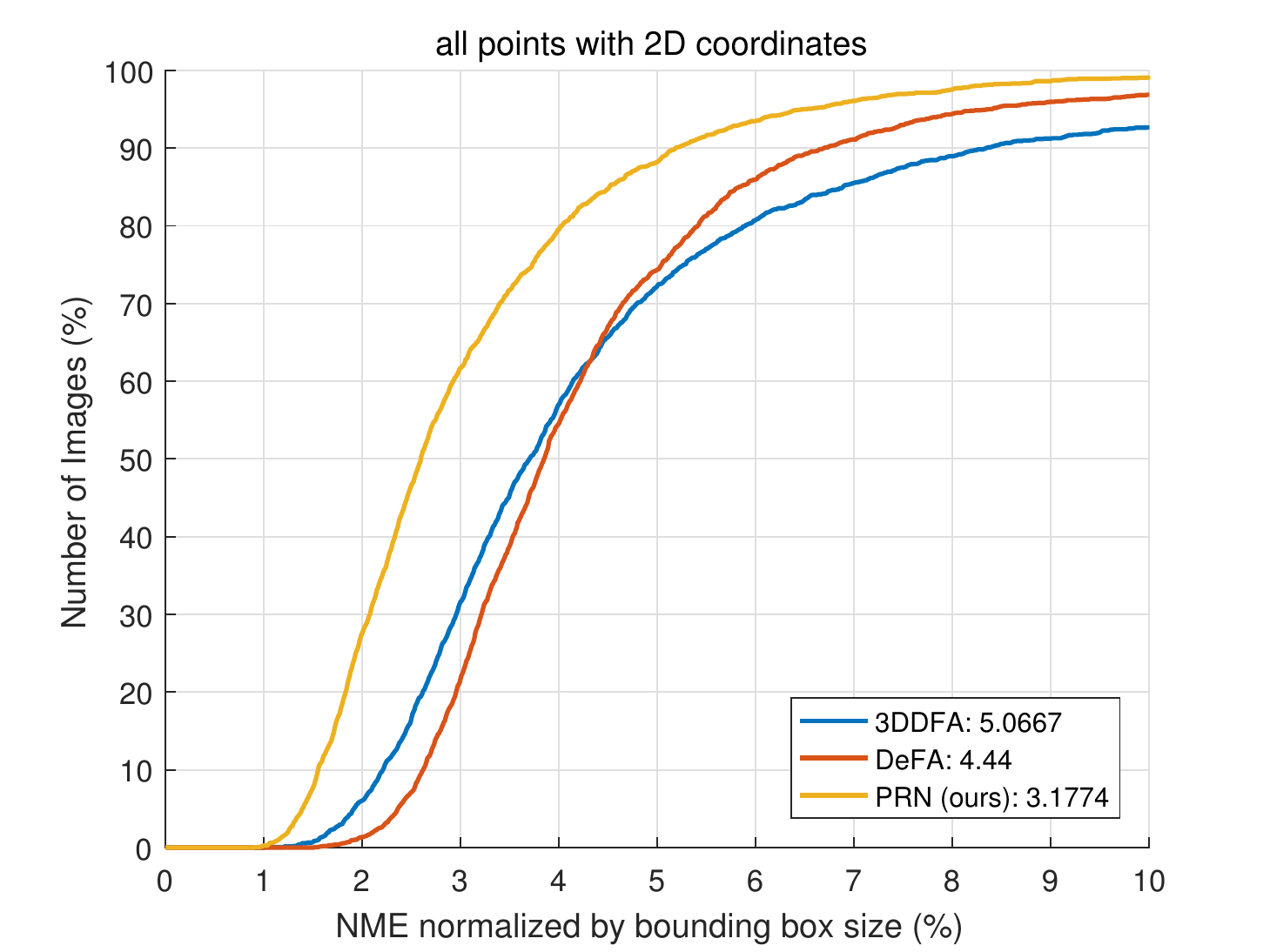}}
  \subfloat{
    \includegraphics[width=0.45\linewidth]{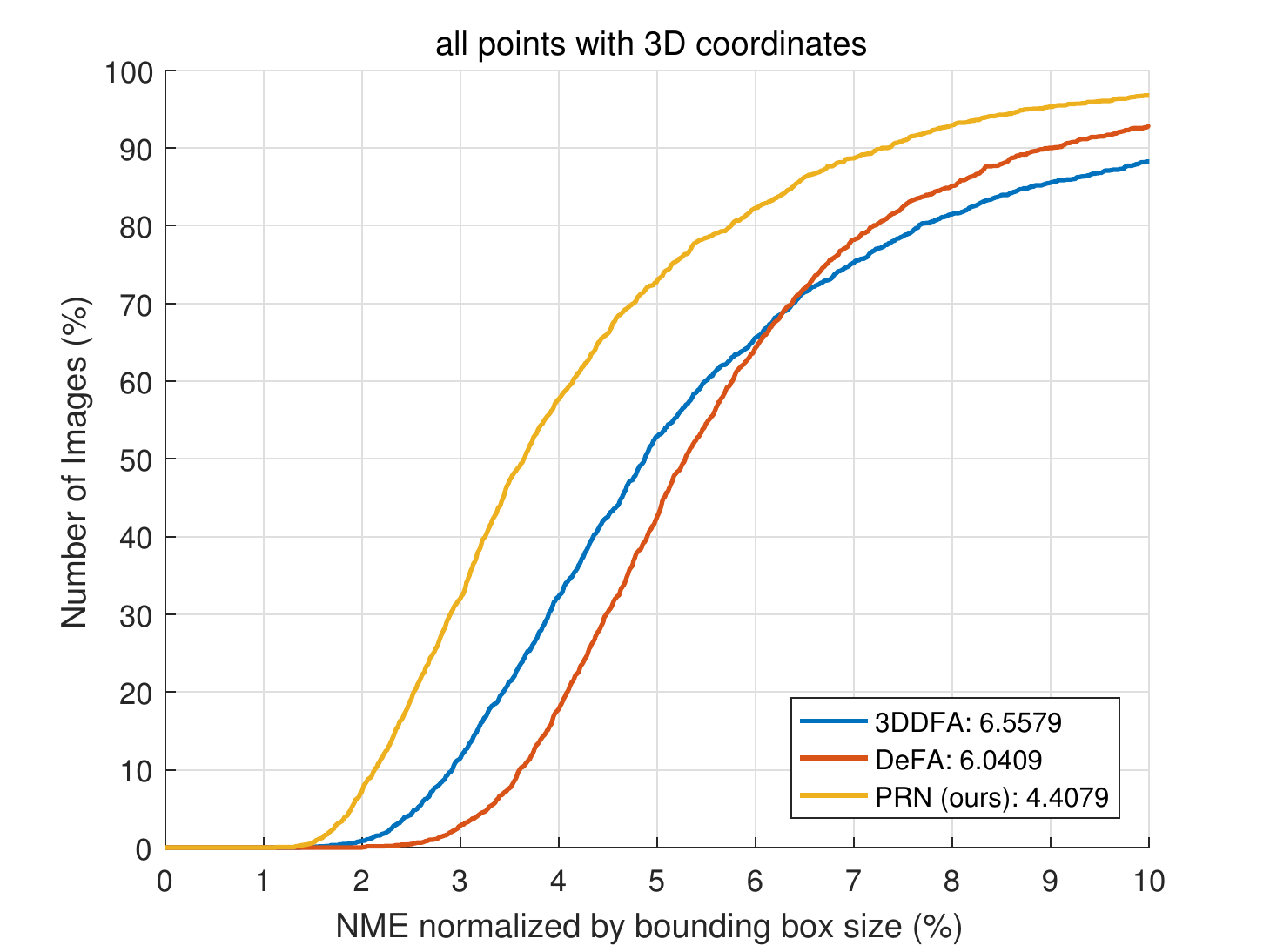}}
  \caption{CED curves on AFLW2000-3D. Evaluation is performed on all points with both the 2D (left) and 3D (right) coordinates. Overall 2000 images from AFLW2000-3D dataset are used here. The mean NME\%  of each method is also showed in the legend.}
  \label{fig:aflw2000 dense} %% label for entire figure
  \vspace{-6mm}
\end{figure}

% Evaluation task: dense alignment(AFLW2000)
We also compare our dense alignment result against other state-of-the-art methods including 3DDFA\cite{zhu2016face} and DeFA\cite{liu2017dense} on the only test dataset AFLW2000-3D. 
In order to compare different methods with the same set of points, we select the points from the largest common face region provided by all methods, and finally around 45K points were used for the evaluation. 
As shown in figure \ref{fig:aflw2000 dense}, our method outperforms the best methods with a large margin of more than \textbf{27\%} on both 2D and 3D coordinates.

% Evaluation settings(metric)
%The classical evaluation metric on 68 landmarks in AFLW2000-3D dataset is the MSE normalized by the ground truth bounding box size, specifically, the size is the square root of the box enclosing the given landmarks.
%Previous methods only calculate the distance with 2D coordinates, which is not real 3D. Thus, we also evaluate the NME with depth value. 
%Specifically, the mean value of depth differs due to the difference of methods, and we align the predicted depth with the ground truth according to its mean value. 
%On the task of dense alignment, AFLW2000-3D dataset offers 3DMM coefficients that can generate the corresponding 3D mesh. Thus, we use the same way in our training step to generate the 256x256 point cloud, and evaluate the only face region points which used in our method. Then for better comparison, we use the same metric as in evaluating 68 landmarks, we compare our method with 3DDFA and DeFA on the evaluation of 45K points from the ground truth 3D face mesh. 
%In addition, the bounding box is also used in our experiments to crop the images. Particularly, we try our best to crop the images as the public code in 3DDFA, DeFA, 3D-FAN sets to re-implement the results described in their published paper. 
%Note that, the codes for evaluating 3D-FAN is the python version the author provided. 

\subsection{3D Face Reconstruction} 
\label{sec: reconstruction}
In this part, we evaluate our method on 3D face reconstruction task and compare with 3DDFA\cite{zhu2016face}, DeFA\cite{liu2017dense} and VRN-Guided\cite{Jackson2017Large} on AFLW2000-3D and Florence datasets. 
We use the same set of points as in evaluating dense alignment and changes the metric so as to keep consistency with other 3D face reconstruction evaluation methods. 
We first use Iterative Closest Points(ICP) algorithm to find the corresponding nearest points between the network output and ground truth point cloud, then calculate Mean Squared Error(MSE) normalized by outer interocular distance of 3D coordinates. 
% use Mean Squared Error(MSE) normalized by outer interocular distance of 3D points here, and Iterative Closest Points(ICP) algorithm is used to find the corresponding nearest points between the network output and ground truth point cloud. 
%In this part, we conduct evaluation in the task of 3D face reconstruction. The comparison approaches are two model-based methods 3DDFA and DeFA, one end-to-end method VRN-Guided\cite{Jackson2017Large}. The evaluation metric is the same as in \cite{Jackson2017Large}: the MSE normalized by outer interocular distance calculated with 3D points.

\begin{figure}
  \vspace{-3mm}
  \centering
  \subfloat{
    \includegraphics[width=0.45\linewidth]{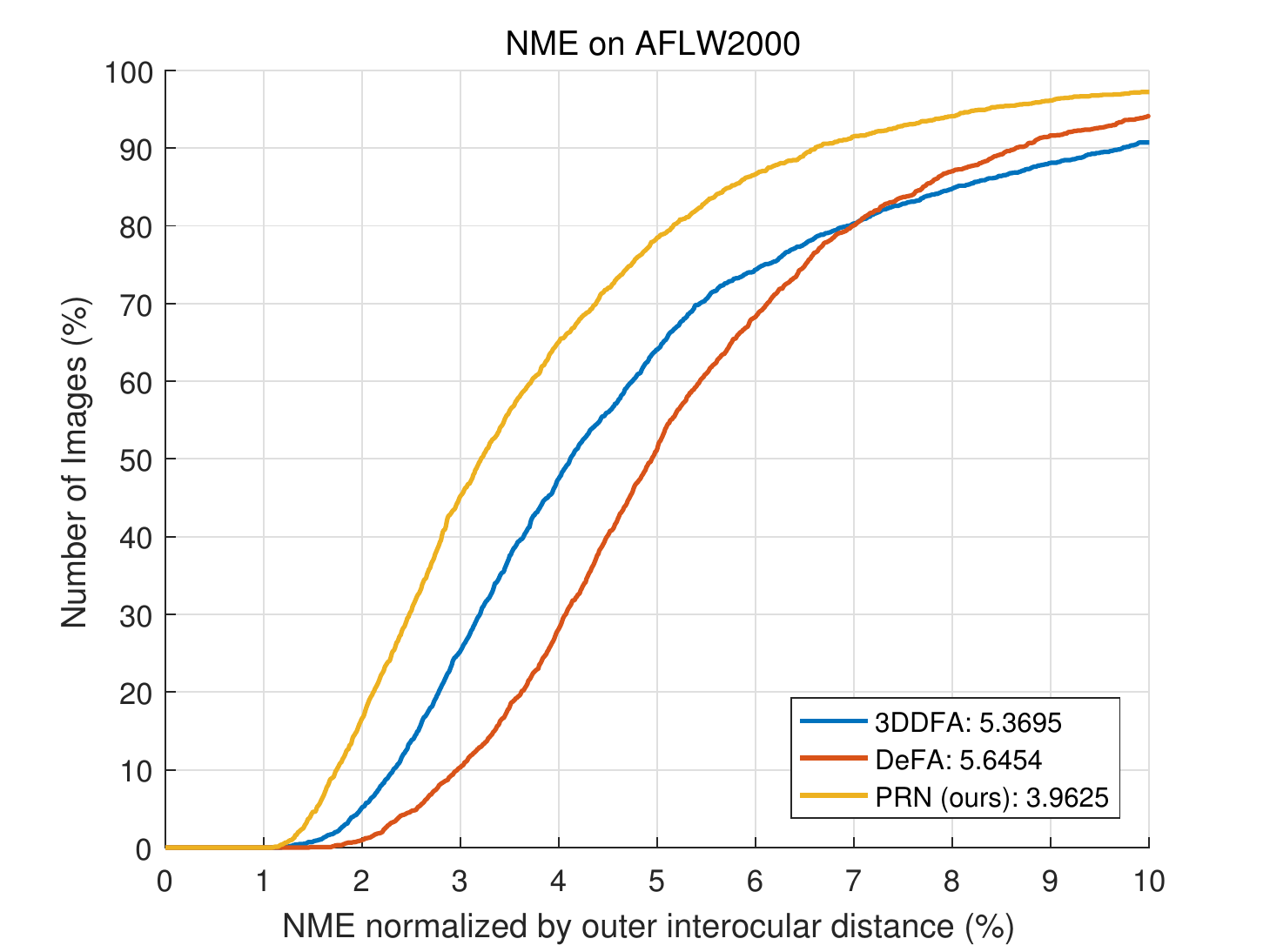}}
  \subfloat{
    \includegraphics[width=0.45\linewidth]{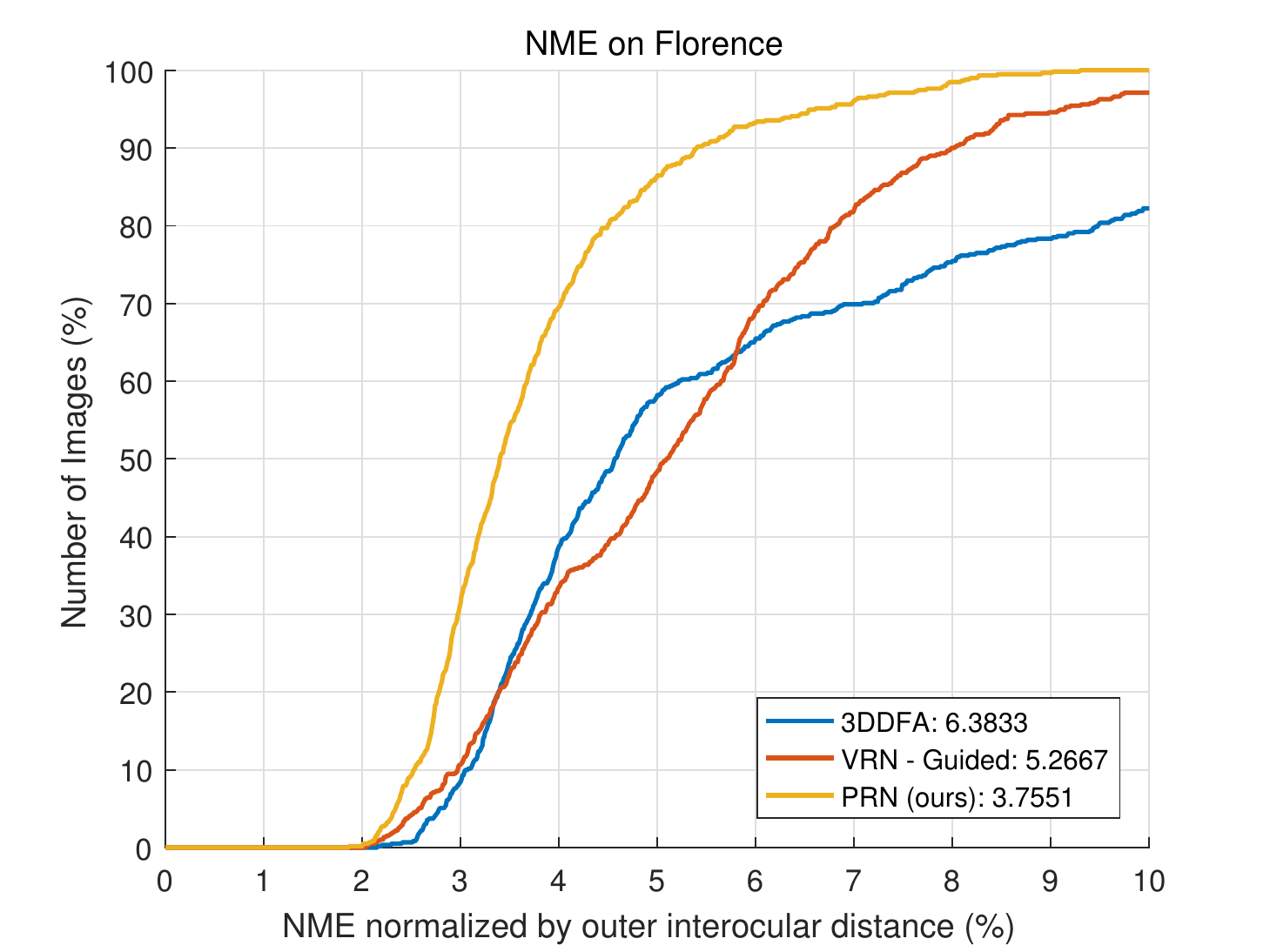}}
  \caption{3D reconstruction performance(CED curves) on in-the-wild AFLW2000-3D dataset and Florence dataset. The mean NME\%  of each method is showed in the legend. On AFLW2000-3D, more than 45K points are used for evaluation. On Florence, about 19K points are used. }
  \label{fig: reconstruction} 
  \vspace{-3mm}
\end{figure}

The result is shown in figure \ref{fig: reconstruction}. our method greatly exceeds the performance of other two state-of-the-art methods.
%On test dataset AFLW2000-3D, we compare our method with 3DDFA and DeFA on the evaluation of 4W points from the ground truth 3D face mesh. Note that, the evaluation is different from the dense alignment, ICP is used here to find other correspondences first. Then the per-pixel distance are calculated. We can see in Figure \ref{fig: reconstruction}, our method largely exceed other two state-of-the-art methods.
% Florence
Since AFLW2000-3D dataset is labeled with results from 3DMM fitting, we further evaluate the performance of our method on Florence dataset, where ground truth 3D point cloud is obtained from structured-light 3D scanning system.
Here we compare our method with 3DDFA and VRN-Guided\cite{Jackson2017Large}, using experimental settings in \cite{Jackson2017Large}. The evaluation images are the renderings with different poses from Florence database, we calculate the bounding box from the ground truth point cloud and using the cropped image as network input. 
%We compare our methods against VRN-Guided on this dataset, and the experimental settings are the same with \cite{Jackson2017Large}. The evaluation images are the renderings with different poses from Florence database, we calculate the bounding box from the ground truth point cloud and using the cropped image as our input. 
Although our method output more complete face point clouds than VRN, we only choose the common face region to compare the performance, 19K points are used for the evaluation.
%Our method obtain more than 45K points of face surface, to better evaluate the performance discrepancy between VRN-Guided and our method, we only use ICP to find the corresponding 19K points. Note that, we only align z value coarsely, no rigid transformation like rotation is conducted.
Figure \ref{fig: reconstruction} shows that our method achieves \textbf{28.7\%} relative higher performance compared to VRN-Guided on Florence dataset, which is a significant improvement.

\begin{figure}
  \vspace{-3mm}
  \centering
  \subfloat{
    \includegraphics[width=0.42\linewidth]{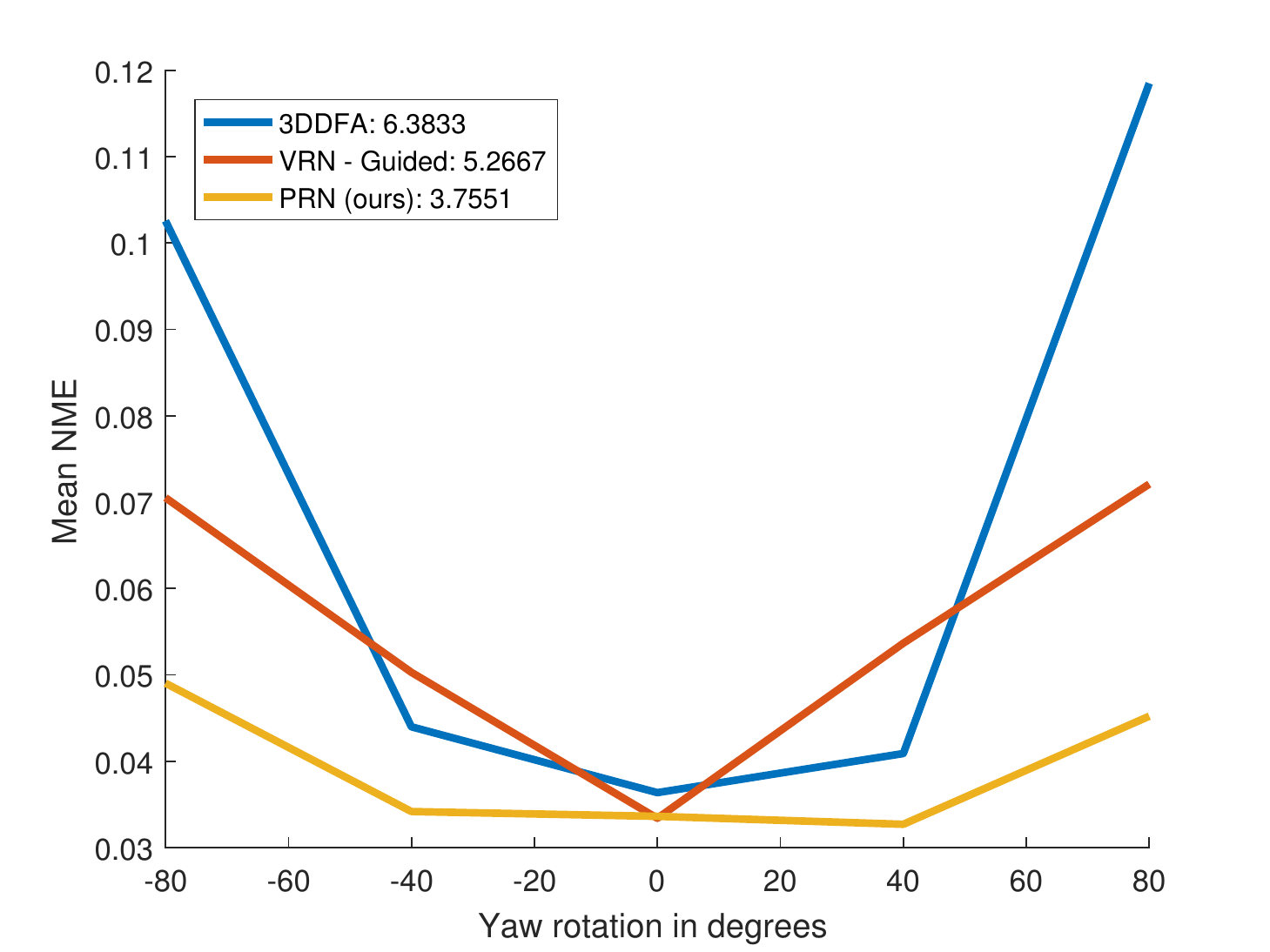}}
%   \hspace{0.1in}
  \subfloat{
    \includegraphics[width=0.46\linewidth]{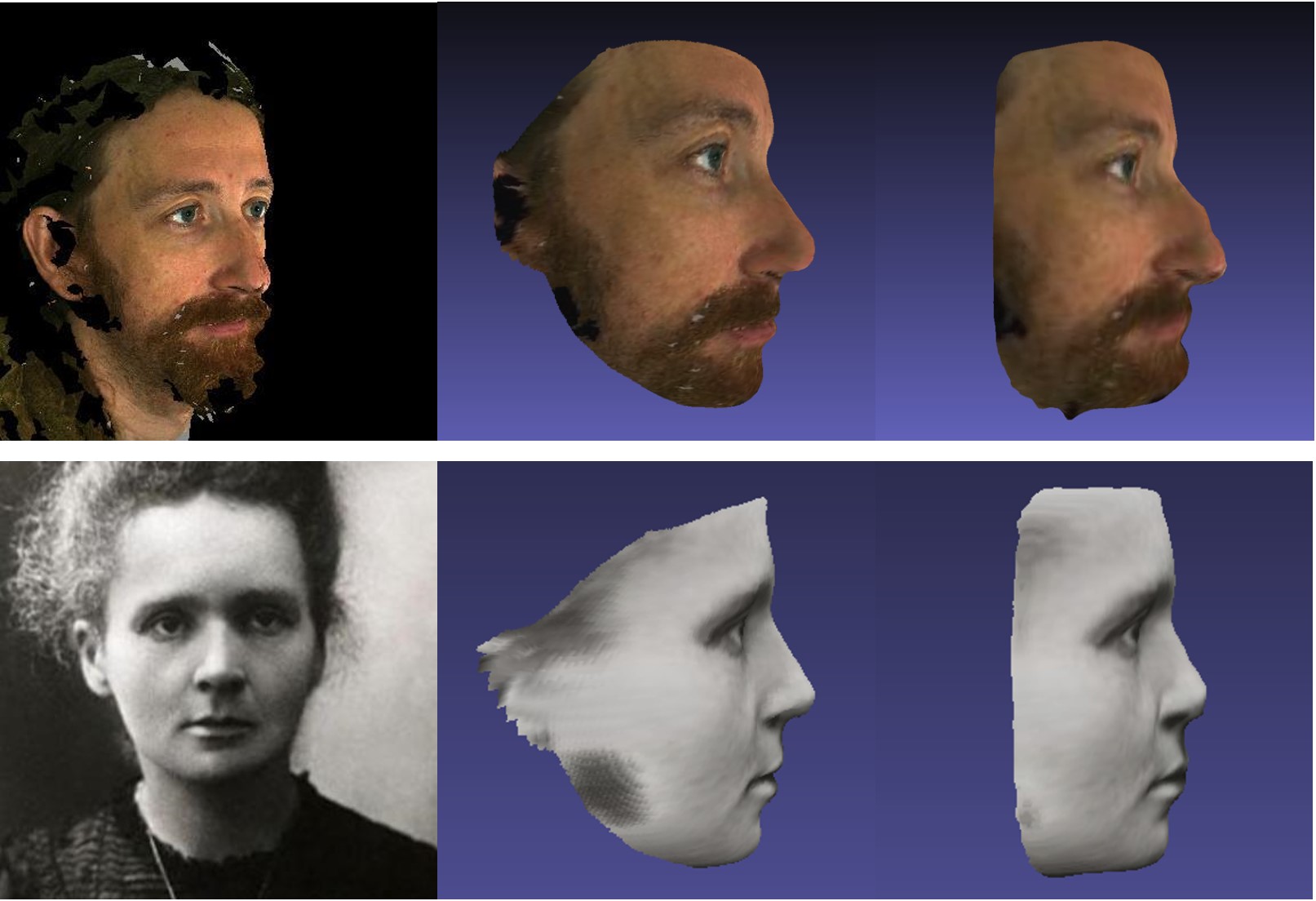}}
      %\hspace{1in}
  \caption{Left: CED curves on Florence dataset with different yaw angles. Right: the qualitative comparison with VRN-Guided. The first column is the input images from Florence dataset and the Internet, the second column is the reconstructed face from our method, the third column is the results from VRN-Guided.}
  \label{fig: comparison} %% label for entire figure
  \vspace{-3mm}
\end{figure}

% poses
To better evaluate the reconstruction performance of our method across different poses, we calculated the NME for different yaw angle range. As shown in figure \ref{fig: comparison}, all the methods perform well in near frontal view, however, 3DDFA and VRN-Guided fail to keep low error as pose becomes large, while our method keeps relatively stable performance in all pose ranges. 
%Furthermore, we also investigate our performance across different poses on Florence dataset. The accuracy both decrease along with the increase of poses. However, in front faces the performance of two methods have little difference, yet our methods perform largely than VRN-Guided in profile faces.The results shows out method is more robust to reconstruct faces across poses.
We also illustrate the qualitative comparison in figure \ref{fig: comparison}, our restored point cloud covers a larger region than in VRN-Guided, which ignores the lateral facial parts. 
Besides, due to the limitation on resolution of VRN, our method provides finer details of face, especially on the nose and mouth region.
%Besides, the reconstructed details of our method including nose and mouth are more meticulous than VRN-Guided.

We also provide additional qualitative results on 300VW\cite{chrysos2015offline} and Multi-PIE\cite{Hartley2003Multiple} datasets, please refer to supplementary material for full details.

\subsection{Runtime}
\label{sec: runtime}

Surpassing the performance of all other state-of-the-art methods on 3D face alignment and reconstruction, our method is surprisingly more light-weighted and faster. 
Since our network uses basic encoder-decoder structure, our model size is only 160MB compared to 1.5GB in VRN\cite{Jackson2017Large}.
We also compare the runtime between our method and other state-of-the-art methods. Table \ref{tab: runtime} shows the result.  
The results of 3DDFA and 3DSTN are directly recorded from their published papers and others are recorded by running their publicly available source codes. 
Notice that, We measure the run time of the process which is defined from inputing the cropped face image until recovering the 3D geometry(point cloud, mesh or voxel data) for 3D reconstruction methods or obtaining the 3D landmarks for alignment methods. The harware used for evaluation is an NVIDIA GeForce GTX 1080 GPU and an Intel(R) Xeon(R) CPU E5-2640 v4 @ 2.40GHz.

\begin{table}
\vspace{-3mm}
\begin{center}
\caption{Run time in Milliseconds per Image}
\label{tab: runtime}
\begin{tabular}{|c|c|c|c|c|c|}
\hline
3DDFA\cite{zhu2016face} &  DeFA\cite{liu2017dense} & 3D-FAN\cite{Bulat2017How} & 
3DSTN\cite{bhagavatula2017faster} & VRN-Guided\cite{Jackson2017Large} & PRN (ours)\\
\hline
75.7 & 35.4 & 54.7 & 
19.0 & 69.0 & 9.8\\
\hline
\end{tabular}
\end{center}
\vspace{-3mm}
\end{table}

Specifically, 
DeFA needs 11.8ms(GPU) to predict 3DMM parameters and another 23.6ms(CPU) to generate mesh data from predicted parameters,
3DFAN needs 29.1ms(GPU) to estimate 2D coordinates first and 25.6ms(GPU) to obtain depth value, 
VRN-Guided detects 68 2D landmarks with 28.4ms(GPU), then regress the voxel data with 40.6ms(GPU), 
our method provides both 3D reconstruction and dense alignment result from cropped image in one pass in 9.8ms(GPU).
%Since our 3D representation is a variation of 3D mesh data and regressed without any intermediate information, we only spend time on network inference phase with only 9.8ms on GPU.

\subsection{Ablation Study} 
\label{sec: ablation}
\begin{figure}
  \vspace{-3mm}
  \centering
  \subfloat{
    \includegraphics[width=0.45\linewidth]{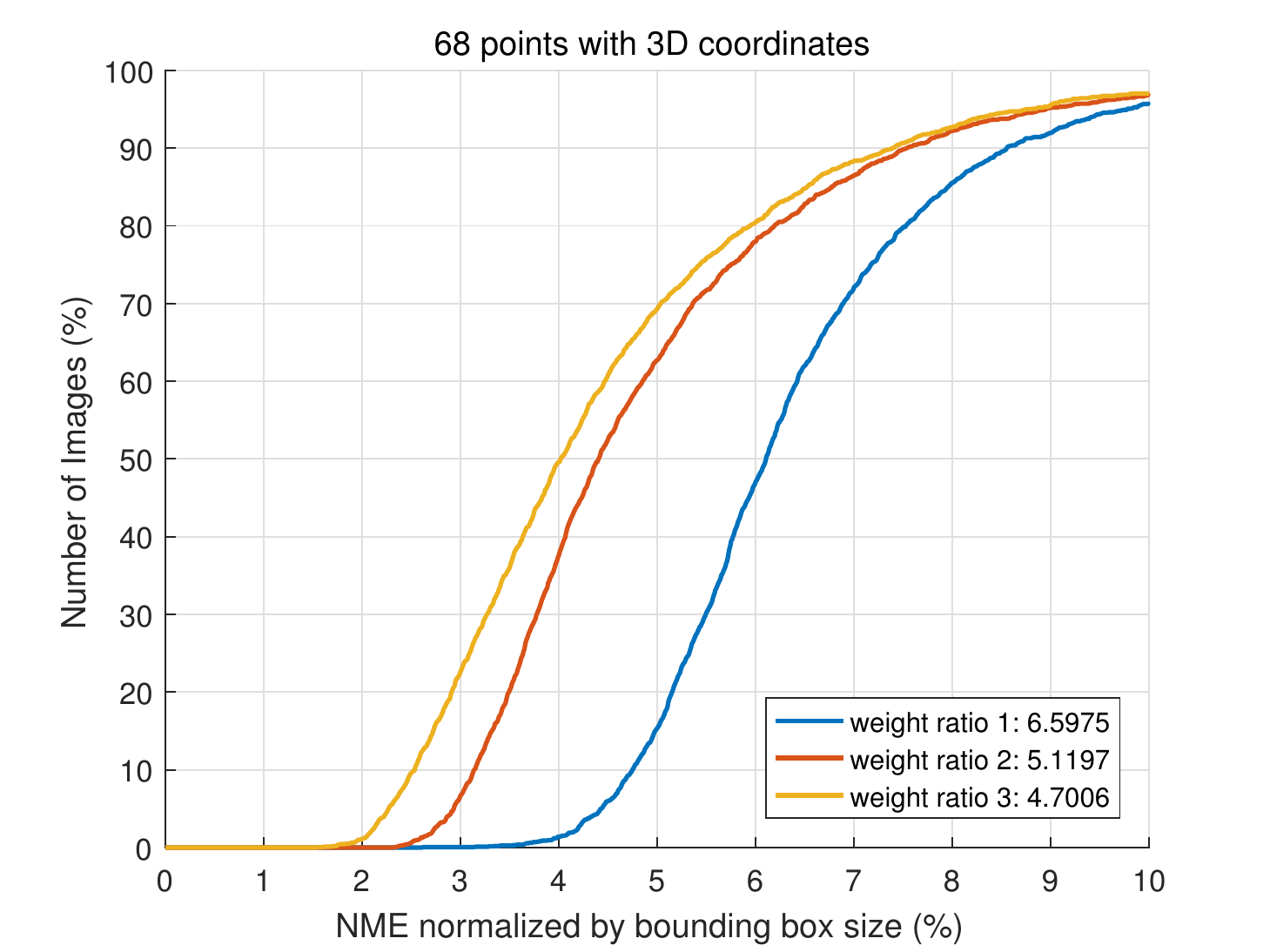}}
  %\hspace{1in}
  \subfloat{
    \includegraphics[width=0.45\linewidth]{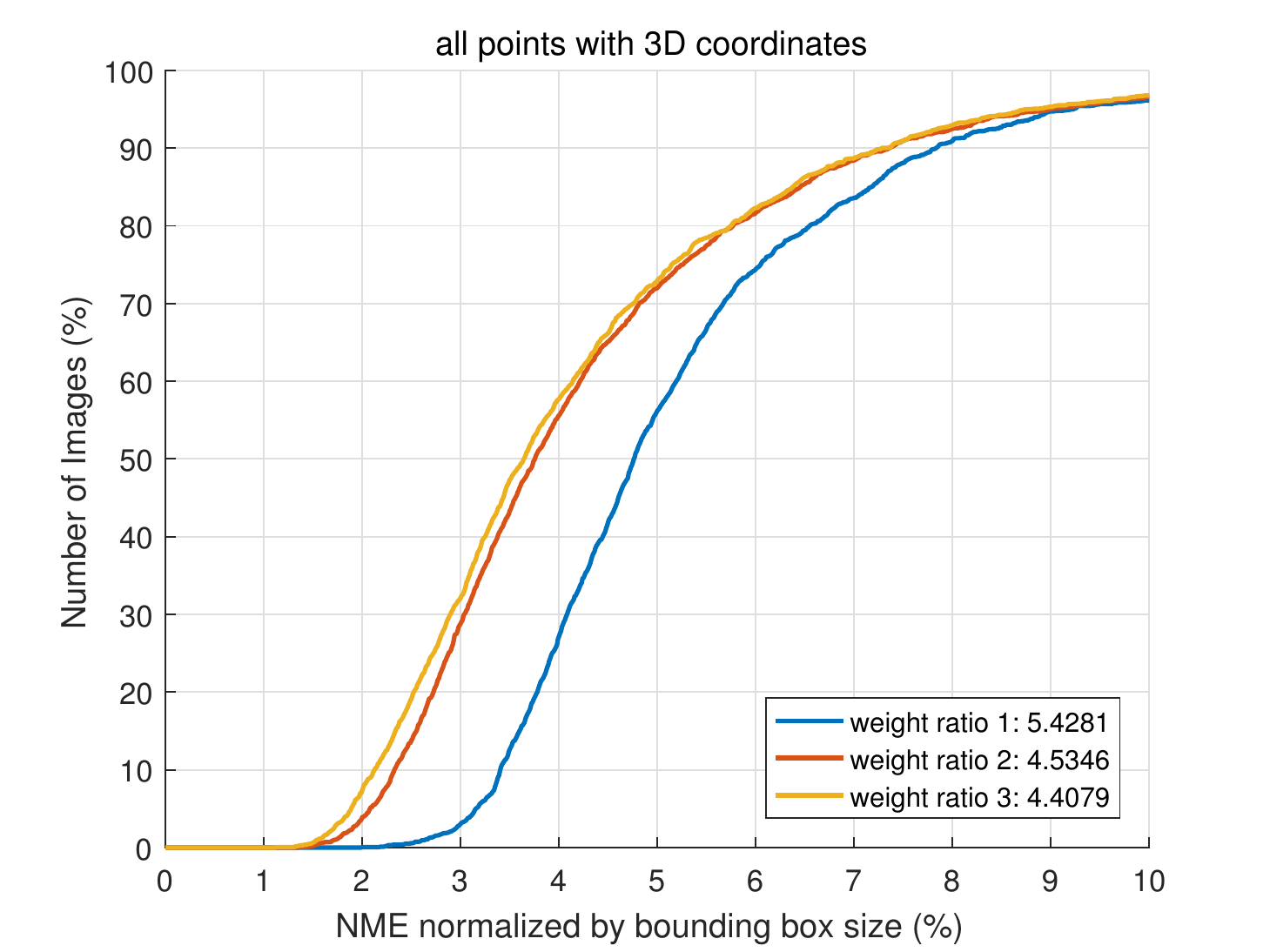}}
      %\hspace{1in}
\caption{The effect of weight mask evaluated on AFLW2000-3D dataset with 68 landmarks(left) and all points(right).}
  \label{fig: ablation} %% label for entire figure
  \vspace{-3mm}
\end{figure}

In this section, we conduct several experiments to evaluate the influence of our weight mask on training and provide both sparse and dense alignment CED on AFLW2000 to evaluate different settings. Table \ref{tab: ratio} shows three options for our weight mask, we could see that weight ratio 1 corresponds to the situation when no weight mask is used, weight ratio 2 and 3 are slightly different on the emphasis in loss function.  
\begin{table}
\vspace{-3mm}
\begin{center}
\caption{The weight ratio of different settings}
\label{tab: ratio}
\begin{tabular}{|c|c|c|c|}
\hline
  &  weight ratio 1 & weight ratio 2 & weight ratio 3 \\
\hline
sub1:sub2:sub3:sub4 & 1:1:1:1 & 1:1:1:0 & 16:4:3:0\\
\hline
\end{tabular}
\end{center}
\vspace{-3mm}
\end{table}
% Comp: with weight mask & without weight 
% We trained the network with different losses and compared the NME for key points alignment and reconstruction. 
The results are shown in Figure \ref{fig: ablation}. Network trained without using weight mask has worst performance compared with other two settings. By adding weights to specific regions such as 68 facial landmarks or central face region, weight ratio 3 shows considerable improvement on 68 points datasets over weight ratio 2.
%Obviously, the usage of weight mask in training step will improve the location of landmarks largely, besides, the weight mask also helps the network to learn discriminative features on the face regions which also help the regression of all 3D points.

\section{Conclusion}
In this paper, we propose an end-to-end method, which well solves the problems of 3D face alignment and 3D face reconstruction simultaneously. By learning the position map, we directly regress the complete 3D structure along with semantic meaning from a single image. Quantitative and qualitative results demonstrate our method is robust to poses, illuminations and occlusions. 
Experiments on three test datasets show that our method achieves significant improvements over others. 
We further show that our method runs faster than other methods and is suitable for real time usage.
% We believe our work can be easily applied to various facial related tasks such as pose estimation, face animation or 3D face fusion.
%Furthermore, the achievement in both alignment and reconstruction tasks make it easy to apply our method to many other face-related tasks such as 3D pose estimation, face animation and 3D face fusion. 
% The test codes and these related-task codes are publicly available at \url{https://github.com/Anonymous7005/PRN}.

\section{Acknowledgements}
We would like to thank Aaron Jackson for kindly providing the Florence test data. We also thank Iacopo Masi for his patience in helping me acquire  Florence 3D Face dataset. This work was supported by CloudWalk Technology.

%\clearpage

\bibliographystyle{splncs}
\bibliography{egbib}
\end{document}